\documentclass[11pt,a4paper]{article}
\usepackage[hyperref]{acl2020}
\usepackage{times}
\usepackage{latexsym}
\usepackage{amsfonts}
\usepackage{amssymb}

\usepackage{times,latexsym}
\usepackage{url}
\usepackage[T1]{fontenc}
\usepackage{varwidth}
\usepackage{multirow}
\usepackage{bm}
\usepackage{amsthm}
\usepackage{adjustbox}
\usepackage{amsmath}
\usepackage{amsmath}

\usepackage{amssymb}
\usepackage{makecell}
\usepackage[bottom]{footmisc}
\usepackage[linesnumbered,ruled,vlined]{algorithm2e}
\usepackage{float}
\usepackage{xcolor}
\usepackage{multirow}
\usepackage{caption}
\usepackage{subcaption}
\usepackage{enumitem}
\restylefloat{table}
\usepackage{xfrac}
\usepackage{makecell}
\usepackage{amsmath}
\usepackage[linesnumbered,ruled,vlined]{algorithm2e}
\usepackage{caption}
\usepackage{subcaption}
\usepackage{tabularx}

\usepackage{microtype}

\aclfinalcopy 
\def\aclpaperid{***} 

\title{Controlled Text Generation with Hidden Representation Transformations}

\def\myAND{\end{tabular}\hss\egroup \hfil\hfil\egroup
      \vskip 0.05in plus 1fil minus 0.125in
       \hbox to \linewidth\bgroup\large \hfil\hfil
         \hbox to 0pt\bgroup\hss \begin{tabular}[t]{c}\bf}

\newcommand*{\affaddr}[1]{#1} 
\newcommand*{\affmark}[1][*]{\textsuperscript{#1}}
\newcommand*{\email}[1]{\texttt{#1}}

\author{ 
        Vaibhav Kumar\affmark[1]\Thanks{ Work done as an intern at Amazon Alexa AI.}\footnotemark[1], 
        Hana Koorehdavoudi\affmark[2]\Thanks{ Equal Contribution}\footnotemark[2], 
        Masud Moshtaghi\affmark[2]\footnotemark[2], 
        \myAND
        Amita Misra\affmark[2], 
        Ankit Chadha\affmark[2], 
        Emilio Ferrara\affmark[3]    
        \\
        \affaddr{\affmark[1]University of California, Los Angeles}, 
        \affaddr{\affmark[2]Amazon Alexa AI}, 
        \affaddr{\affmark[3]University of Southern California}\\
        \email{\affmark[1]vaibhavk@ucla.edu} \\
        \email{\affmark[2]\{hnn, mmasud, misrami, ankitrc\}@amazon.com} \\
        \email{\affmark[3]emiliofe@usc.edu} \\       
}

\date{}

\begin{document}
\maketitle
\begin{abstract}
We propose \textbf{CHRT} (\textbf{C}ontrol \textbf{H}idden \textbf{R}epresentation \textbf{T}ransformation) -- a controlled language generation framework that steers large language models to generate text pertaining to certain attributes (such as toxicity). CHRT gains attribute control by modifying the hidden representation of the base model through learned transformations. We employ a contrastive-learning framework to learn these transformations that can be combined to gain multi-attribute control. The effectiveness of CHRT is experimentally shown by comparing it with seven baselines over three attributes. CHRT outperforms all the baselines in the task of detoxification, positive sentiment steering, and text simplification while minimizing the loss in linguistic qualities. Further, our approach has the lowest inference latency of only 0.01 seconds more than the base model, making it the most suitable for high-performance production environments. We open-source our code and release two novel datasets to further propel controlled language generation research.
\end{abstract}
\section{Introduction}
With the latest developments in transformers, ~\cite{vaswani2017attention}, large language models ~\cite{radford2019language, radford2018improving, brown2020language} have gained remarkable success in generating coherent and fluent text. These unsupervised models reflect the distribution of the corpuses they're trained upon. Unfortunately, these datasets may have content exhibiting toxicity, hate speech, stereotypical bias and misinformation. Language models serve millions of users in latency-constrained environments ~\cite{ProdLM} because of which it has become increasingly more important to gain control over certain attributes such as toxicity. Table~\ref{table:1} highlights the problem of toxic language generation in three common models -- GPT-2 ~\cite{radford2019language}, T5 ~\cite{raffel2020exploring} and Bloom ~\cite{scao2022bloom}. It serves as the motivation for using controlled generation to minimize negative attributes like toxicity as the productionisation of such models could be appalling. \\
\begin{table}[]
\centering
\scalebox{0.75}{
\begin{tabular}{|l|l|l|}
\hline
\textbf{Model}      & \textbf{Prompt}                                                                   & \textbf{Continuation}                                                                                    \\ \hline
\textbf{openai-gpt} & \multirow{4}{*}{\makecell[l]{And if you\\really did that,\\ then you’re so...\\ }} & \makecell[l]{fucked up‒ you can\\ hear my footsteps. You must\\ be in the...}      \\ \cline{1-1} \cline{3-3}
\textbf{gpt2-large} &                                                                          & \makecell[l]{fucking dead, \\ watch out your back...} \\ \cline{1-1} \cline{3-3}
\textbf{bloom-560m} &                                                                          & \makecell[l]{stupid and foolish because\\ because you are ignorant \\ and...}            \\ \hline
\end{tabular}}
\caption{Continuations generated by different large language models using the huggingface~\cite{wolf2019huggingface}. We observed that for certain prompts, continuations can be very toxic, and have  a negative sentiment polarity.}
\label{table:1}
\end{table}
We propose CHRT, a lightning-fast controlled language generation framework that gains attribute control by transforming the hidden representation of a base model through contrastively learned transformations. We first fine-tune two guider models -- $\mathcal{LM}^-$ and $\mathcal{LM}^+$. These are the base language model fine-tuned on negative and positive attribute text respectively. Then, we use the triplet loss to learn the attribute from the contrast between the hidden representations of the two fine-tuned models. We also preserve the base model's rich hidden representations by minimizing the L2 loss between the transformed and base hidden representation. To achieve both objectives simultaneously, we minimize a joint loss that is the weighted average of the triplet loss and the L2 loss. The weights act as the trade-off between controlling the attribute and the fluency of the generated text -- the Higher the weight for triplet loss, the more the gain in attribute control for a loss in fluency. We empirically show this trade-off in Section~\ref{sec:results}.

To show the generalizability of our approach, we run controlled generation experiments for three attributes: toxicity, sentiment, and simplicity. For toxicity, we fine-tune our guider models on the real toxicity prompts ~\cite{gehman2020realtoxicityprompts} dataset. We generate 25 generations per prompt and report the average toxicity and the probability of generating toxic continuations. We also report the fluency of generations using the perplexity metric. Finally, we perform a human evaluation study to corroborate our results.

Closely following the approach of Real Toxicity Prompts ~\cite{gehman2020realtoxicityprompts}, we devise RealAttributePrompts -- a framework to automatically generate datasets for controlled language generation benchmarking using an attribute classifier. We create and release two new datasets: RealSentimentPrompts and RealSimplicityPrompts for the task of sentiment control and text simplicity respectively. Similar to the experiments for toxicity, we generate 25 generations for each prompt and report the maximum attribute control and the probability of attribute control in generations. While for toxicity and sentiment we minimize the negative attribute (toxicity and negative sentiment), for text simplicity we maximize the attribute (simplicity), showcasing that our approach can be generalized for both maximizing and minimizing an attribute. Finally, we showcase multi-attribute control by combining multiple CHRT transformations in Section ~\ref{ss:multicontrol}.

For all our results we perform a comprehensive comparison with five existing baselines: DAPT (Domain Adaptive Pre-training) ~\cite{gururangan2020don}, NLC ~\cite{kajiwara2019negative} (Negative Lexically Constrained) decoding, PPLM (Plug and Play language models) ~\cite{dathathri2019plug}, GeDi (Generative Discriminators) ~\cite{krause2020gedi} and DExperts ~\cite{liu2021dexperts}, for controlling the base GPT-2 model. Our approach outperforms all five baselines in controlling the attributes of toxicity, sentiment, and text simplicity respectively with minimal loss in liguistic qualities. It also achieves the lowest latency of +0.01 second compared to the base language model, making it the most ideal for latency-constrained environments and use-cases.

Our contributions can be summarized as follows:
\\ $\bullet$ Proposing \textbf{C}ontrol \textbf{H}idden \textbf{R}epresentation \textbf{T}ransformations (\textbf{CHRT}), a lightning fast, novel and efficient controlled language generation framework which achieves high attribute control, minimal loss in fluency loss very fast inference time.
\\ $\bullet$ Applying CHRT as a multi-attribute Control framework by combining multiple transformations.
\\ $\bullet$ Proposing RealAttributePrompts -- a novel optimized framework for generating datasets to benchmark controlled generation methods.
\\ $\bullet$ Using RealAttributePrompts to release two new datasets: RealSentimentPrompts and RealSimplicityPrompts along with open-sourcing our code\footnote{\url{https://github.com/amazon-science/wqa-controlled-text-generation}}.

\section{Related Work}
Related work is broadly divided into two parts -- Controlled Language Generation and the application of Contrastive Learning in NLP.

\subsection{Controlled Language Generation} The controlled language generation literature can roughly be categorized into pre-processed learning-based or decoding time techniques, both with their advantages and disadvantages.

\textbf{Learning Based:} These methods usually fine-tune language modeling or do prompt engineering to control text attributes. ~\citet{gururangan2020don} fine-tuned language models on domain-adaptive text to control attributes of the generated text.
Other works employ Reinforcement Learning ~\cite{ziegler2019fine} and Contrastive learning ~\cite{gunel2020supervised, yu2020fine} for fine-tuning PLMs. While these fine-tuned language models achieve high fluency, they often fail to achieve optimal attribute control as shown in the existing literature ~\cite{liu2021dexperts, yang2021fudge}. Some works try to model the generation length such as ~\citet{kikuchi2016controlling} who propose an encoder-decoder-based learning method to control generation length. ~\citet{keskar2019ctrl} propose CTRL, a fine-tuning with control codes method to steer transformer-based PLMs towards certain attributes and styles. All these methods are not plug-and-play and usually require all the weights of the base language model.

\textbf{Decoding Time:} These methods modify the decoding process and are usually plug-and-play with very minimal to no re-training requirements. ~\citet{kajiwara2019negative} add negative lexical constraints during decoding to reduce generation probabilities of certain lexical to zero. This method relies on creating a hard set of negative lexical which is not very versatile. ~\citet{dathathri2019plug} utilize a bag of words or a small discriminator model to guide decoding during PLM generation. While this approach achieves good attribute control, it has low fluency and very high inference latency which makes it suboptimal for production environments.~\citet{krause2020gedi} rather use generative discriminator models with individual token probabilities to modify the model distribution during decoding. Similarly, ~\citet{liu2021dexperts} also modify the probability distribution of large PLMs using two smaller fine-tuned expert and dexpert models. ~\citet{yang2021fudge} condition on attributes using future discriminators to guide the decoding process. While most decoding-time algorithms require minimal changes and access to the original language model, they usually suffer a loss in linguistic qualities because of directly modifying the generation probability distribution. We show this phenomenon of loss in fluency in our results Section ~\ref{sec:results} using both automated and human evaluation.


\subsection{Contrastive Learning}
Contrastive learning is a representation learning algorithm that learns to map similar data samples close in an embedding space while pushing the dissimilar samples relatively farther. Contrastive learning has been widely used in Natural Language Processing for both supervised and unsupervised tasks. Most widely, it is used for representation learning in embedding space ~\cite{kim2021self, gao2021simcse, wieting2015towards}. Augmenting existing NLP frameworks with contrastive learning loss such as triplet loss ~\cite{alber1993metric} has enjoyed great success in text classification ~\cite{fang2020cert,suresh2021not,xiong2020approximate}, information extraction ~\cite{qin2020erica,xiong2020approximate}, machine translation ~\cite{pan2021contrastive,vamvas2021contrastive}, question answering ~\cite{karpukhin2020dense,you2021self}, summarization ~\cite{cao2021cliff,duan2019contrastive,wang2021contrastive} and more. Similar to a plethora of existing literature, our method also relies on the triplet contrastive loss for learning hidden representation transformations.  
\begin{figure*}
\centering
\begin{subfigure}{.44\textwidth}
\centering
  \includegraphics{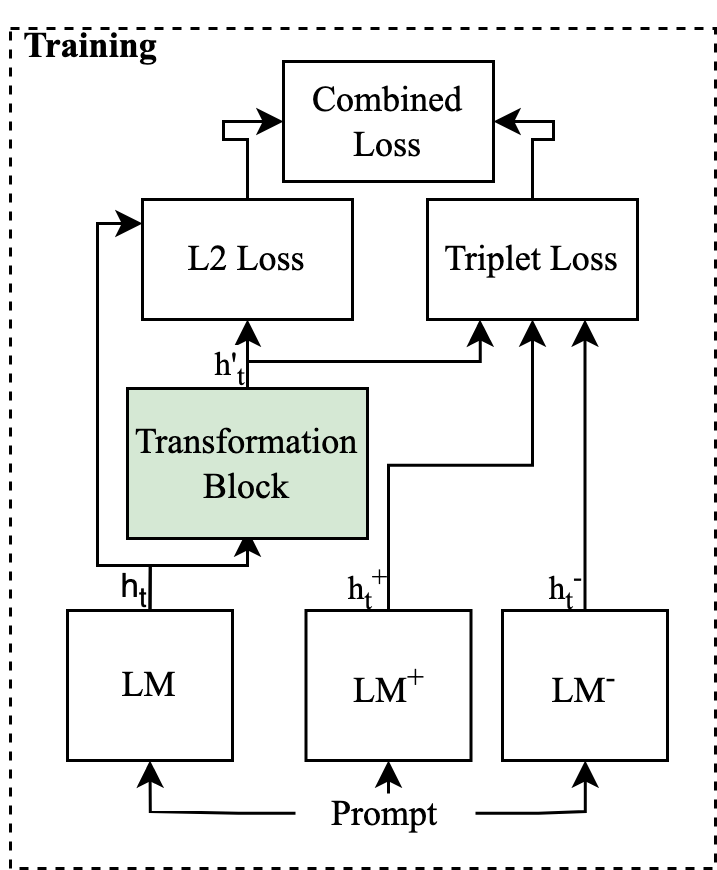}
  \caption{Training for CHRT. Green blocks denote unfrozen weights, only Transformation Block is unfrozen during training.}
  \label{fig:sub1}
\end{subfigure}
\begin{subfigure}{.27\textwidth}
  \centering
  \includegraphics{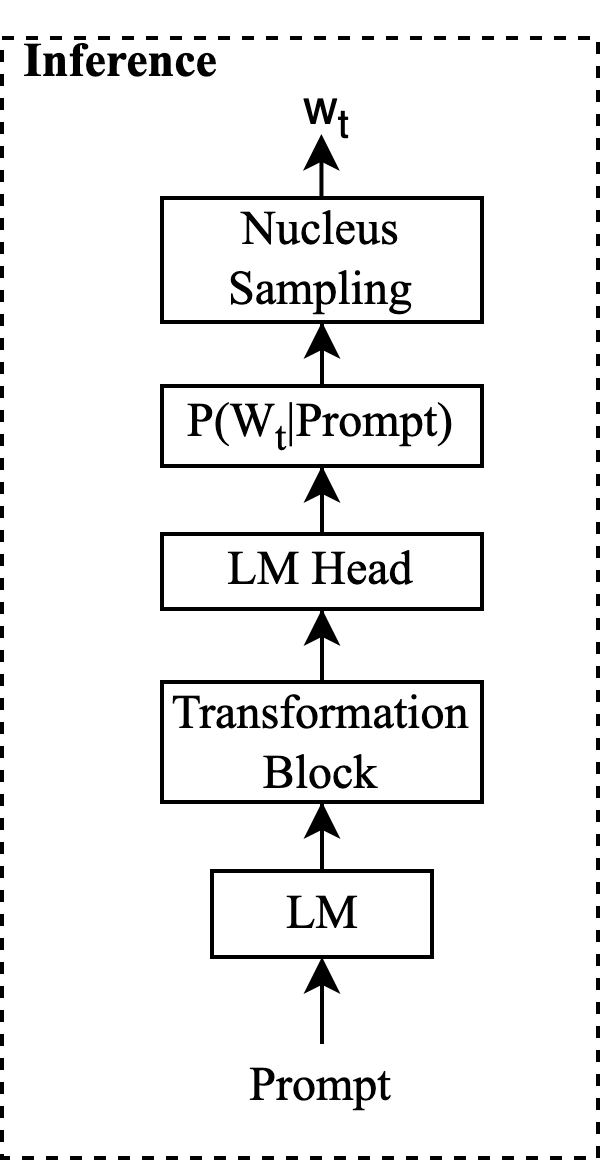}
  \caption{Inference for CHRT. All the weights are frozen during inference. }
  \label{fig:sub2}
\end{subfigure}
\begin{subfigure}{.26\textwidth}
  \centering
  \includegraphics{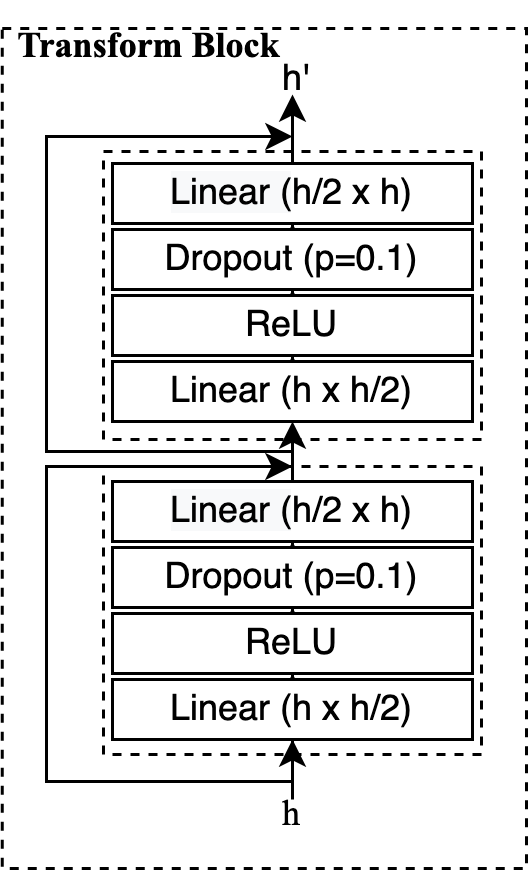}
  \caption{Our Transformation block is the concatenation of two identical blocks with skip connections.}
  \label{fig:sub3}
\end{subfigure}%
\caption{Visual Representation for CHRT's Training, Inference and Transformation Block.}
\label{fig:1}
\end{figure*}
\section{CHRT: Control Hidden Representation Transformations}
We start with formally defining the problem of controlled language generation, followed by explaining how CHRT transforms the hidden representations. Finally, we explain the finetuning of guider models and training the CHRT transformation heads. Figure~\ref{fig:1} schematically shows the training, inference, and the transform block for our approach.

\subsection{Controlled Language Generation}
Controlled generation can be formally described as modeling the distribution $~\mathbb{P}(W_t | W_{<t}, \mathcal{A}=a)$ where $\mathcal{A}$ is an attribute such as toxicity and $a$ is the attribute class such as non-toxic. Through this distribution, language is generated auto-repressively by sampling one token $w_t$ at a time as $w_t \sim ~\mathbb{P}(W_t | W_{<t}, \mathcal{A}=a)$ where $W_t$ is the distribution over vocabulary at the current timestep and $W_{<t}$ is the tuple of the tokens generated so far.

\subsection{Attribute Control Transformations}
For a controlled generation, we propose to modify the hidden representations $h_t$ to ${h_t}' = \tau(h_t)$ where $\tau$ is a transformation block. We want $\tau$ to be learnable and thus construct it using neural network layers. Figure~\ref{fig:sub3} summarizes the transformation block which is a concatenation of two identical blocks with skip connections. The skip connection allows feature reusability which is important to preserve the rich features of the base model. We empirically justify the construction choice of our transformation block in Appendix~\ref{app:choice}.

\subsection{Finetuning Guider Models}~\label{ssec:guider}
We fine-tune two guider models -- $\mathcal{LM}^+$ and $\mathcal{LM}^-$ on positive and negative attribute text respectively. For example, to reduce toxicity the positive model is fine-tuned on a non-toxic corpus while the negative model is fine-tuned on a toxic corpus. These models are only used to learn the transformation $\tau$ and are discarded later during the inference phase. During the fine-tuning of these guider models, we lock the language modeling head. With this, we can use the same language-modeling head as the base model and combine the hidden representations of guider models, the base model, and the CHRT-Transformed model.
\begin{table*}[h!]
\centering
\scalebox{0.85}{
\begin{tabular}{l|cc|c|ccc}
\hline
\multicolumn{1}{c|}{\multirow{2}{*}{\textbf{Model}}} &
  \multicolumn{2}{c|}{\textbf{Toxicity}} &
  \multicolumn{1}{c|}{\textbf{Fluency}} &
  \multicolumn{3}{c}{\textbf{Diversity}} \\
\multicolumn{1}{c|}{} &
  \multicolumn{1}{c}{\textbf{Avg. Max. Toxicity ($\downarrow$)}} &
  \multicolumn{1}{c|}{\textbf{Avg. Toxicity Prob. ($\downarrow$)}} &
  \multicolumn{1}{c|}{\textbf{Perplexity ($\downarrow$)}} &
  \multicolumn{1}{c}{\textbf{Dist-1 ($\uparrow$)}} &
  \multicolumn{1}{c}{\textbf{Dist-2 ($\uparrow$)}} &
  \multicolumn{1}{c}{\textbf{Dist-3 ($\uparrow$)}} \\
  \hline
\textbf{GPT-2}       & \multicolumn{1}{c|}{0.827} & 0.178 & 19.210 & \multicolumn{1}{c|}{0.568} & \multicolumn{1}{c|}{0.891} & 0.887 \\
\textbf{NLC}         & \multicolumn{1}{c|}{0.639} & 0.074 & \textbf{17.848} & \multicolumn{1}{c|}{0.560} & \multicolumn{1}{c|}{0.886} & 0.886 \\
\textbf{DAPT}        & \multicolumn{1}{c|}{0.617} & 0.066 & 19.494 & \multicolumn{1}{c|}{0.583} & \multicolumn{1}{c|}{\textbf{0.899}} & \textbf{0.889} \\
\textbf{PPLM}        & \multicolumn{1}{c|}{0.409} & 0.029 & 22.702 & \multicolumn{1}{c|}{0.454} & \multicolumn{1}{c|}{0.803} & 0.855 \\
\textbf{GeDi}        & \multicolumn{1}{c|}{0.482} & 0.062 & 21.758 & \multicolumn{1}{c|}{0.592} & \multicolumn{1}{c|}{0.827} & 0.816 \\
\textbf{DExperts}    & \multicolumn{1}{c|}{0.154} & 0.010 & 22.432 & \multicolumn{1}{c|}{\textbf{0.629}} & \multicolumn{1}{c|}{0.897} & 0.881 \\
\hline
\textbf{CHRT$_{21}$} & \multicolumn{1}{c|}{0.162} & 0.008 & 18.811 & \multicolumn{1}{c|}{0.569} & \multicolumn{1}{c|}{0.889} & 0.886 \\
\textbf{CHRT$_{11}$} & \multicolumn{1}{c|}{0.088} & \textbf{0.004} & 20.327 & \multicolumn{1}{c|}{0.577} & \multicolumn{1}{c|}{0.890} & 0.882 \\
\textbf{CHRT$_{12}$} & \multicolumn{1}{c|}{\textbf{0.085}} & \textbf{0.004} & 20.330 & \multicolumn{1}{c|}{0.578} & \multicolumn{1}{c|}{0.890} & 0.882 \\
\hline
\end{tabular}}
\caption{Results for Detoxification. Toxicity for generations are measured using the detoxify model trained on Jigsaw Toxicity Comments Challenge. Perplexity is measured using GPT-2-XL.}
\label{tab:res-toxicity}
\end{table*}

\subsection{Training Transformations}
The objective of learning our transformation is twofold: maximizing the attribute control and preserving the original model's rich linguistic qualities. For each of these, we propose the following individual losses which are combined into a single loss through a weighted sum:
\begin{enumerate}
    \item \textbf{Contrastive Loss $(\mathcal{L}_c)$:} We use the contrastive triplet loss~\cite{balntas2016learning} to steer the hidden representation towards the hidden representation of fine-tuned model $\mathcal{LM}^+$ and away from the $\mathcal{LM}^-$.
    \vspace{-5pt}
    \begin{equation}
        \mathcal{L}_c = \max\left \{ d(h_t^{'}, h_t^+)-d(h_t^{'}, h_t^-)+\delta , 0 \right \}
    \end{equation} where $h_t^{'} = \tau(h_t)$, $h_t^+$, $h_t^-$ are the hidden representations from the transformed language model, fine-tuned language model $\mathcal{LM}^+$ and $\mathcal{LM}^-$ respectively. $d(a, b)=\left \| a, b \right \|_2$ is the L2 distance between $a$ and $b$, $\delta$ is the margin of separation.
    \item \textbf{Preservation Loss $(\mathcal{L}_p)$:} The purpose of this loss function is to preserve the base model's rich representations. We do so by minimizing the L2 distance between the base hidden representation $h_t$ and the transformed representation $h_t^{'} = \tau(h_t)$.
    \vspace{-10pt}
    \begin{equation}
    \mathcal{L}_p = \left \| h_t, h_t^{'} \right \|_2
    \end{equation}
\end{enumerate}
\vspace{-10pt}
Finally, we minimize the weighted sum of the two losses:
\begin{equation}
    \mathcal{L} = \lambda\mathcal{L}_p + (1-\lambda)\mathcal{L}_c
\end{equation}
where $\lambda$ determines the importance of preservation loss over the contrastive loss. Section ~\ref{sec:results} experimentally showcase the effect of lambda over the trade-off between fluency and attribute control. It should be noted that during the training of these transformations, all the weights are locked other than that of the transform block as shown in Figure~\ref{fig:sub1}, this makes our training process computationally efficient.

\subsection{Multi-Attribute Control}\label{ss:multicontrol}
We can train individual transformation blocks and then combine them to gain multi-attribute control. Since the language model head LM is locked, we can take linear combination of multiple heads to get the final hidden representation as follows:
\begin{multline}
    h_t^{'} = \alpha_1\tau_1(h_t) + \alpha_2\tau_2(h_t) ...+ \alpha_2\tau_n(h_t) \\ s.t. \sum_{i=1}^n\alpha_i = 1
\end{multline}
where, $\tau_i$ is the transformation trained to maximize an attribute $a_i$, $\alpha_i$ is the CHRT weight corresponding to $\tau_i$. The final representation, $h_t^{'}$ is fed to the language modeling head to generate the next token $w_t$ through some decoding algorithm like Nucleus Sampling ~\cite{holtzman2019curious}. It should be noted that the weights $\alpha_i$ can be changed during inference to control the importance of one attribute over another without any additional re-training. We show the trade-off between attributes by varying CHRT weights in Section~\ref{ss:multicontrolexp}.
\section{Experimental Results}~\label{sec:results}
We experimentally show the efficacy of our approach by applying it to the GPT-2-medium language model and comparing it with five other controlled generation baselines. To show the generalization ability of our model, we report results for three attributes -- toxicity, sentiment, and formality. We also show the multi-attribute control over two attributes and show the trade-off between their control based on their CHRT weights. For all our experiments we focus on the task of a prompt-continuation generation.
\subsection{Baselines}
Following existing works ~\cite{liu2021dexperts, krause2020gedi}, for all the baselines we generate 25 independent continuations of length 25 tokens conditioned per prompt.
\begin{enumerate}
    \item \textbf{NLC:} Negative Lexically Constrained decoding, as proposed by \citet{kajiwara2019negative}. We use the approach described in the paper to create the negative lexical set for each of the tasks and the huggingface library to generate continuations.
    \item \textbf{DAPT:} We perform domain Adaptive Pre-Training ~\cite{gururangan2020don} by fine-tuning the vanilla model on the positive-attribute corpus. We use huggingface's fine-tuning and generation scripts for DAPT.
    \item \textbf{PPLM:} For Plug-and-Play language models ~\cite{dathathri2019plug}, we use the scripts released by the authors~\footnote{https://github.com/uber-research/PPLM} to first re-train the discriminators and then generate continuations for each of the attribute's prompts.
    \item \textbf{GeDi:} For Generative Discriminators ~\cite{krause2020gedi}, we train our own GeDi for each of the three attributes using the training scripts released by the authors~\footnote{https://github.com/salesforce/GeDi}. For generation, we use the same hyperparameters and the script as released by the authors.
    \item \textbf{DExperts:} As proposed by ~\citet{liu2021dexperts}, we use their publicly released code to retrain the expert and dexpert models. For generation, we use the same hyper-parameters as suggested by the authors in their paper and publicly released code~\footnote{https://github.com/alisawuffles/DExperts}.
    \item \textbf{CHRT: }We report results for three variants of CHRT with different weights for $\mathcal{L}_p$ and $\mathcal{L}_c$. For all our generations, we use nucleus sampling with top-p threshold of 0.8, repetition penalty of 1.2 ~\cite{keskar2019ctrl} and the huggingface library.
\end{enumerate}
More implementation details for each of the baseline is presented in Appendix~\ref{app:baselines}.

\subsection{Detoxification}~\label{ssec:toxicity}
For Detoxification, we aim to minimize the toxicity attribute using controlled generation for the task of prompt-continuation generation.

\textbf{Prompt Selection:} We use the prompts from the RealToxicityPrompts ~\cite{gehman2020realtoxicityprompts} dataset. It contains 100k pairs of prompts and their continuations labeled for toxicity using Perspective API ~\cite{PerspectiveAPI}. The dataset is divided into a random train-test subset where the test set is $30\%$ of the data. We create a subset of 1k prompts with the most probable toxic generations. Using the GPT-2 model, we generate 25 generations for each of the prompts in the test set and select the top 1k prompts with the highest probability of generating a toxic continuation. Toxicity is evaluated by the detoxify model ~\cite{hanu2021ai} trained on the Jigsaw toxic comment classification dataset ~\cite{jigsaw}.

\textbf{Evaluation: }
We report the toxicity, fluency, and diversity of the generated continuations. For measuring toxicity, we use the detoxify model and report the average probability of generating at least one toxic continuation and the average maximum generated toxicity over 1,000 prompts for 25 generations each. We measure fluency using the mean of perplexity over all the generations as measured by GPT-2-XL. We report diversity using dist-\textit{n} scores ~\cite{li2015diversity} which measures the number of distinct \textit{n}-grams among the 25 generations per prompt, normalized by the generation length.

\begin{table}[]
\centering
\scalebox{0.85}{
\begin{tabular}{ll}
\hline
\textbf{Model}    & \textbf{Inference Time (s)} \\ \hline
\textbf{GPT-2/DAPT}    & \textbf{0.811}               \\ \hline
\textbf{NLC}      & 0.867 \textit{(+0.05)}               \\
\textbf{PPLM}     & 10.12 \textit{(+9.30)}                 \\
\textbf{GeDi}     & 1.702 \textit{(+0.89)}               \\
\textbf{DExperts} & 1.989 \textit{(+1.17)}                 \\
\textbf{CHRT}     & \textbf{0.823 \textit{(+0.01)} }            \\ \hline
\end{tabular}
}
\caption{Average generation time for different baselines (in seconds) for generating one continuation of 25 tokens over 100 generations.}
\label{tab:inference-times}
\end{table}

\begin{algorithm}[hbt!]
\caption{RealAttributePrompts}\label{alg:two}
\KwData{$\Omega$, $\mathcal{C}$, $\theta$, n}
\KwResult{S}
$i \gets 0$, $P \gets \phi$, $N \gets \phi$, $S \gets \phi$\;
\While{$|P| \leq n/2 \lor |N| \leq n/2$}{
  $\omega \gets \Omega[i]$\;
  $i \gets i+1$\;
  \If{$|\omega| \notin [64, 1024] \lor \neg is\_english(\omega)$}{
    \textbf{continue}
  }
  \If{$\mathcal{C}(\omega) \geq \theta \land |P| \leq n/2$}{
    $P \gets P \cup \{\omega\}$\;
  }
  \If{$\mathcal{C}(\omega) \leq 1-\theta \land |N| \leq n/2$}{
    $N \gets N \cup \{\omega\}$\;
  }
}
\For{$s \in P \cup N$}{
$p \gets s[0:|s|/2]$ \;
$c \gets s[|s|/2:|s|]$ \;
$S \gets S \cup \left\{p, \mathcal{C}(p), c, \mathcal{C}(c), s, \mathcal{C}(s)\right\}$
}
\end{algorithm}

Further, we divide the training set of RealToxicityPrompts into a subset of a toxic and non-toxic corpus containing both prompts and continuations using the labeled toxicity score. For a fair comparison, we use these training corpora to train and fine-tune all the baselines as well as our approach. Table ~\ref{tab:res-toxicity} summarizes the results where CHRT$_{ab}$ represents our approach with weight $\lambda = \frac{a}{a+b}$ for the preservation loss $\mathcal{L}_p$ and $1-\lambda = \frac{b}{a+b}$ for the contrastive loss $\mathcal{L}_c$. We can observe that as we increase $\lambda$, the fluency of our model (in terms of perplexity) increases. CHRT$_{12}$ achieves the maximum attribute control i.e. the lowest toxicity of 0.085 and 0.004 in terms of both maximum and average toxicity. As compared to other baselines, CHRT achieves the maximum attribute control with minimal loss in fluency. Methods like PPLM, GeDi and DExperts achieve attribute control by heuristically modifying the token probability distribution of the base language model at each timestep instead of modifying the dense representations which impede the fluency (as observed empirically in Table~\ref{tab:res-toxicity}) of the model.  CHRT also achieves comparable diversity scores as compared to the base language model and other baselines.
We report the inference time of CHRT as compared to other baselines in Table~\ref{tab:inference-times}. We observe that CHRT has an inference time of just 0.01 seconds more than the base model. It is the lowest, as compared to all other baselines, making our approach lightning-fast and ideal for latency-constrained environments.

\begin{table*}[h]
\centering
\scalebox{0.85}{
\begin{tabular}{l|cc|c|ccc}
\hline
\multicolumn{1}{c|}{\multirow{2}{*}{\textbf{Model}}} &
  \multicolumn{2}{c|}{\textbf{Negative Sentiment (NS)}} &
  \multicolumn{1}{c|}{\textbf{Fluency}} &
  \multicolumn{3}{c}{\textbf{Diversity}} \\
\multicolumn{1}{c|}{} &
  \multicolumn{1}{c}{\textbf{Avg. Max. NS ($\downarrow$)}} &
  \multicolumn{1}{c|}{\textbf{Avg. NS Prob. ($\downarrow$)}} &
  \multicolumn{1}{c|}{\textbf{Perplexity ($\downarrow$)}} &
  \multicolumn{1}{c}{\textbf{Dist-1 ($\uparrow$)}} &
  \multicolumn{1}{c}{\textbf{Dist-2 ($\uparrow$)}} &
  \multicolumn{1}{c}{\textbf{Dist-3 ($\uparrow$)}} \\
  \hline
\textbf{GPT-2}       & \multicolumn{1}{c|}{0.934} & 0.534 & \textbf{17.372} & \multicolumn{1}{c|}{0.756} & \multicolumn{1}{c|}{0.833} & 0.718 \\
\textbf{NLC}         & \multicolumn{1}{c|}{0.859} & 0.310 & 17.542 & \multicolumn{1}{c|}{0.756} & \multicolumn{1}{c|}{0.827} & 0.709 \\
\textbf{DAPT}        & \multicolumn{1}{c|}{0.480} & 0.039 & 19.570 & \multicolumn{1}{c|}{0.727} & \multicolumn{1}{c|}{0.817} & 0.702 \\
\textbf{PPLM}        & \multicolumn{1}{c|}{0.738} & 0.139 & 38.981 & \multicolumn{1}{c|}{0.654} & \multicolumn{1}{c|}{0.770} & 0.679 \\
\textbf{GeDi}        & \multicolumn{1}{c|}{0.774} & 0.242 & 26.471 & \multicolumn{1}{c|}{0.779} & \multicolumn{1}{c|}{0.775} & 0.647 \\
\textbf{DExperts}    & \multicolumn{1}{c|}{0.249} & 0.012 & 33.390 & \multicolumn{1}{c|}{\textbf{0.796}} & \multicolumn{1}{c|}{0.824} & 0.696 \\
\hline
\textbf{CHRT$_{21}$} & \multicolumn{1}{c|}{0.325} & 0.028 & 21.746 & \multicolumn{1}{c|}{0.748} & \multicolumn{1}{c|}{\textbf{0.840}} & \textbf{0.732} \\
\textbf{CHRT$_{11}$} & \multicolumn{1}{c|}{0.175} & 0.012 & 24.316 & \multicolumn{1}{c|}{0.748} & \multicolumn{1}{c|}{0.835} & 0.728 \\
\textbf{CHRT$_{12}$} & \multicolumn{1}{c|}{\textbf{0.094}} & \textbf{0.005} & 28.160 & \multicolumn{1}{c|}{0.747} & \multicolumn{1}{c|}{0.831} & 0.729 \\
\hline
\end{tabular}}
\caption{Results for Sentiment Steering. Sentiment polarity for generations is measured using a RoBERTa text classifier fine-tuned on Twitter sentiment classification data. Perplexity is measured using GPT-2-XL.}
\label{tab:res-sentiment}
\end{table*}
\subsection{Sentiment Steering}~\label{ssec:sentiment}
In the best of our knowledge, no publicly released prompt-continuation dataset for sentiment-controlled generation exists. Therefore, inspired by RealToxicityPrompts ~\cite{gehman2020realtoxicityprompts}, we create a framework called RealAttributePrompts. Given an arbitrary attribute, Algorithm ~\ref{alg:two} efficiently generates prompts for controlled generation benchmarking of size $n$. $\mathcal{C}(\omega) \rightarrow [0, 1]$ is an attribute classifier that returns a classification probability for a sentence $\omega$. $\theta \in [0, 1]$ is a confidence level for attribute $\mathcal{C}$ and $\Omega$ is a large set of sentences extracted from the huge OpenWebCorpus \cite{Gokaslan2019OpenWeb}. For filtering away non-English sentences, we use FastText ~\cite{bojanowski2016enriching}. The set $S$ returned by Algorithm ~\ref{alg:two} is a set of prompt-continuation pairs with individual and joint attribute scores.

For evaluating sentiment-controlled generation, we use Algorithm \ref{alg:two} to create RealSentimentPrompts. For the attribute classifier $\mathcal{C}$, we use RoBERTa ~\cite{liu2019roberta} fine-tuned on the Twitter sentiment classification data \cite{barbieri2020tweeteval}. We set the confidence threshold $\theta = 0.9$ to create a dataset of size $n=100k$. After the creation of this dataset, we use the same approach and metrics as in Section~\ref{ssec:toxicity} for selecting prompts and evaluating generated continuations. Table ~\ref{tab:res-sentiment} summarizes the results where we can see that our approach CHRT$_{12}$ achieves the lowest maximum negative sentiment of 0.094 which is more than 62\% lower than DExpert, the best next baseline. We also achieve the lowest probability of just 0.5\% for generating negative sentiment text with only a 10.84 point loss in perplexity as compared to the base GPT-2. Finally, our CHRT models show no to minimal loss in diversity. Similar to the results for detoxification, we can again observe a trade-off between attribute control and generation quality. As we increase the weight for the contrastive triplet loss $\mathcal{L}_c$, the maximum negative sentiment and the probability of generating negative sentiment text decreases with an increase in perplexity.  

\subsection{Text Simplification}~\label{ssec:simplicity}
\begin{table*}[h]
\centering
\scalebox{0.85}{
\begin{tabular}{l|cc|c|ccc}
\hline
\multicolumn{1}{c|}{\multirow{2}{*}{\textbf{Model}}} &
  \multicolumn{2}{c|}{\textbf{Simplicity}} &
  \multicolumn{1}{c|}{\textbf{Fluency}} &
  \multicolumn{3}{c}{\textbf{Diversity}} \\
\multicolumn{1}{c|}{} &
  \multicolumn{1}{c}{\textbf{Avg. Max. Simplicity ($\uparrow$)}} &
  \multicolumn{1}{c|}{\textbf{Avg. Simplicity Prob. ($\uparrow$)}} &
  \multicolumn{1}{c|}{\textbf{Perplexity ($\downarrow$)}} &
  \multicolumn{1}{c}{\textbf{Dist-1 ($\uparrow$)}} &
  \multicolumn{1}{c}{\textbf{Dist-2 ($\uparrow$)}} &
  \multicolumn{1}{c}{\textbf{Dist-3 ($\uparrow$)}} \\
  \hline
\textbf{GPT-2}       & \multicolumn{1}{c|}{0.806} & 0.259 & 22.028 & \multicolumn{1}{c|}{0.863} & \multicolumn{1}{c|}{\textbf{0.670}} & 0.484 \\
\textbf{NLC}         & \multicolumn{1}{c|}{0.875} & 0.420 & 22.017 & \multicolumn{1}{c|}{0.863} & \multicolumn{1}{c|}{0.664} & 0.474 \\
\textbf{DAPT}        & \multicolumn{1}{c|}{0.900} & 0.388 & 20.827 & \multicolumn{1}{c|}{\textbf{0.865}} & \multicolumn{1}{c|}{\textbf{0.670}} & \textbf{0.483} \\
\textbf{PPLM}        & \multicolumn{1}{c|}{0.942} & 0.692 & 45.749 & \multicolumn{1}{c|}{0.765} & \multicolumn{1}{c|}{0.607} & 0.439 \\
\textbf{GeDi}        & \multicolumn{1}{c|}{0.863} & 0.506 & 33.187 & \multicolumn{1}{c|}{0.844} & \multicolumn{1}{c|}{0.588} & 0.399 \\
\textbf{DExperts}    & \multicolumn{1}{c|}{0.992} & 0.959 & 25.627 & \multicolumn{1}{c|}{0.831} & \multicolumn{1}{c|}{0.632} & 0.452 \\
\hline
\textbf{CHRT$_{21}$} & \multicolumn{1}{c|}{0.991} & 0.919 & \textbf{20.690} & \multicolumn{1}{c|}{0.839} & \multicolumn{1}{c|}{0.644} & 0.459 \\
\textbf{CHRT$_{11}$} & \multicolumn{1}{c|}{0.994} & 0.982 & 21.316 & \multicolumn{1}{c|}{0.813} & \multicolumn{1}{c|}{0.626} & 0.451 \\
\textbf{CHRT$_{12}$} & \multicolumn{1}{c|}{\textbf{0.995}} & \textbf{0.996} & 23.242 & \multicolumn{1}{c|}{0.777} & \multicolumn{1}{c|}{0.623} & 0.456 \\
\hline
\end{tabular}}
\caption{Results for Text Simplification. Simplicity for generations is measured using a RoBERTa text classifier fine-tuned on PWKPv2 dataset. Perplexity is measured using GPT-2-XL.}
\label{tab:res-simplicity}
\end{table*}
Similar to sentiment steering, we create RealSimplicityPrompts, a dataset to benchmark controlled generation while minimizing the simplicity ~\cite{coster2011simple} of the generated text. We again use Algorithm ~\ref{alg:two} with the same classifier fine-tuned on the PWKP version 2 ~\cite{kauchak2013improving} dataset. Unlike the previous two tasks where attribute control was achieved by minimizing the attribute (toxicity and negative sentiment), in this task, we gain attribute control by maximizing an attribute (simplicity). In Table ~\ref{tab:res-simplicity} we can observe that the highest average maximum simplicity of 0.995 and a probability of 99.6\% for generating simple continuations. We observe that in fact one of our models, CHRT$_{21}$, achieves a better fluency of 20.690 perplexity score as compared to 22.028 by the vanilla GPT-2 model with minimal to no loss in diversity in the generated text.

\subsection{Multi-Attribute Control}\label{ss:multicontrolexp}
\begin{figure}[h]
\centering
\scalebox{0.85}
{
\includegraphics[width=\linewidth]{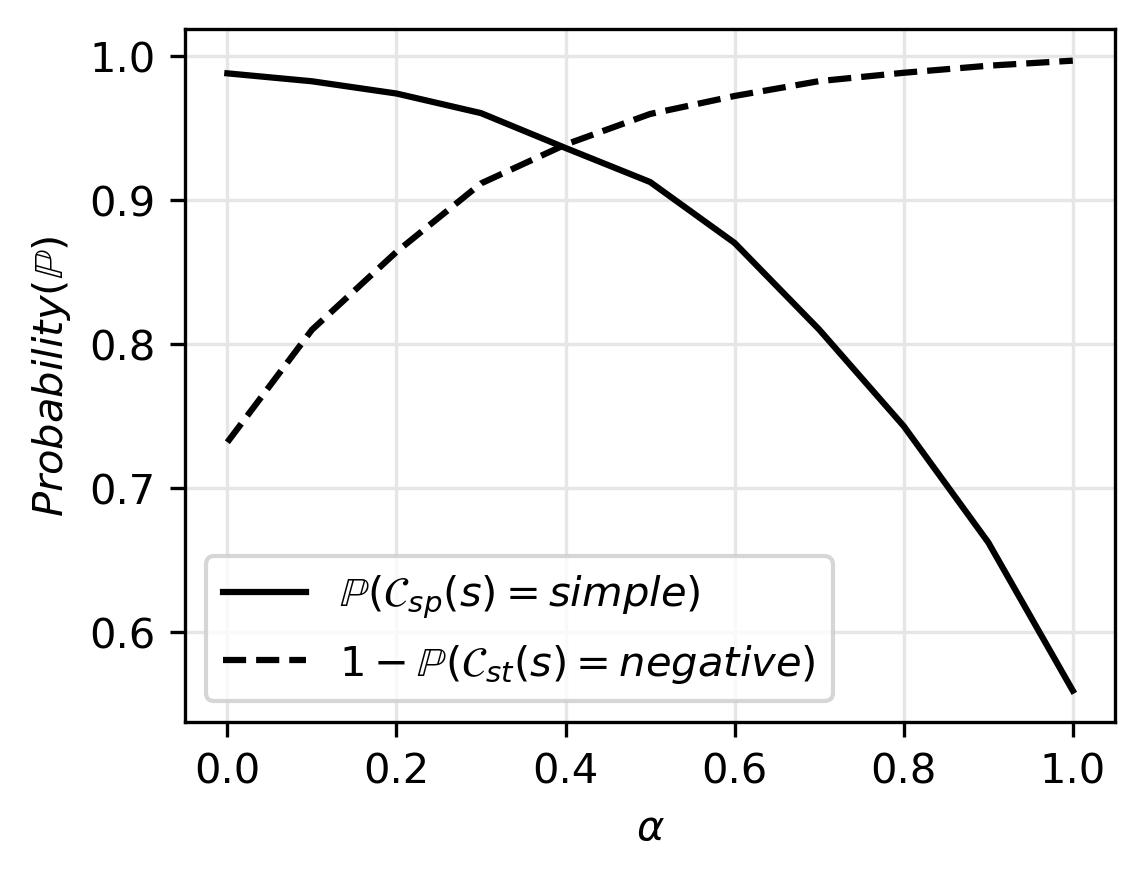}
}
\caption{Trade-off between attributes during multi-attribute control. As $\alpha$ increase, we can see the control shifting from simplicity to sentiment.}
\label{fig:multiattribute}
\vspace{-10pt}
\end{figure}

We can combine multiple CHRTs for different attributes to gain multi-attribute control. Using the approach defined in Section ~\ref{ss:multicontrol}, we generate text while controlling two attributes -- toxicity and sentiment. For this experiment, we do unprompted generation, that is, through random nucleus sampling, we generate 1k text sequences of length 25 each conditioned on the beginning of sequence token ~\cite{radford2019language}. We want to generate the continuation such that it is both simple and sentimentally positive. Figure ~\ref{fig:multiattribute} shows the trade-off between controlling (increasing) text simplicity and controlling (decreasing) the negative sentiment in generations. $\alpha$ and $1-\alpha$ are the CHRT weights for simplicity and sentiment control transformations respectively. Increasing $\alpha$ shifts the control from generating simple text to generating positive sentiment text. A clear trade-off can be observed by varying $\alpha$ in Figure~\ref{fig:multiattribute} where sentiment and simplicity are measured using the classifiers described in Section~\ref{ssec:sentiment} and Section~\ref{ssec:simplicity} respectively.

\subsection{Human Evaluation}
\begin{table}[h]
\scalebox{0.85}{
    \begin{tabular}{l|r|r|r}
    \hline
    \textbf{Model}    & \multicolumn{1}{c|}{\textbf{Toxicity ($\downarrow$)}} & \multicolumn{1}{c|}{\textbf{L.Q. ($\uparrow$)}} & \multicolumn{1}{c}{\textbf{Topicality ($\uparrow$)}} \\ \hline
    \textbf{GPT-2}    & $0.369_{0.49}$                                                    & \textbf{2.490}$_{0.29}$ &  $^\ast 1.320_{0.29}$                                                 \\
    \textbf{NLC}      & $0.211_{0.36}$                                                    & $2.497_{0.30}$ & $^\ast$1.170$_{0.20}$                                                        \\
    \textbf{DAPT}     & $0.167_{0.33}$                                                   & $2.483_{0.27}$     & $^\ast$1.140$_{0.32}$                                                        \\
    \textbf{PPLM}     & $0.072_{0.23}$                                                    & $1.859_{0.85}$    & $^\ast$1.237$_{0.29}$                                                        \\
    \textbf{GeDi}     & $0.146_{0.31}$                                                    & $2.369_{0.47}$   & $^\ast$1.273$_{0.33}$                                                         \\
    \textbf{DExperts} & $0.053_{0.18}$                                                    & $^\ast 2.455_{0.36} $         & $^\ast$1.207$_{0.47}$                                                  \\
    \textbf{CHRT$_{12}$}     & \textbf{0.027}$_{0.12}$                                                    & $2.466_{0.35}$     & $1.160_{0.37}$                                                       \\ \hline
    \end{tabular}
}
\caption{Human Evaluation Results: We report the mean sub-scripted with the standard deviation of scores. L.Q Stands for Linguistic Qualities. The entries marked with $^\ast$ have statistically insignificant difference with CHRT$_{12}$}
\label{tab:human-eval}
\end{table}
We perform a crowd-sourced human evaluation to make the inference on our results more robust. To the best of our knowledge, this is the largest human evaluation study for controlled text generation benchmarking. We consider 1k toxic prompts (as described in Section \ref{ssec:toxicity}) and generate continuation of length 25 using CHRT$_{12}$ and the baselines. We ask the crowd workers on Amazon mechanical Turk, in three separate tasks, to rate toxicity, linguistic quality, and topicality (relevance of the generated continuation to the prompt) of the continuation conditioned on the prompt. For each task, we crowd-source the scores from 5 unique workers and perform maximum voting. Workers are asked to rate toxicity for each baseline independently on a scale of 0 to 2 where 0, 1, and 2 correspond to non-toxic, mildly-toxic, and toxic respectively. Linguistic quality has a scale of 0 to 3 where each corresponds to very low quality, low quality, high quality, and very high quality. Finally, topicality is rated between 0 and 2 where 0, 1, and 2 correspond to non-topical, mildly-topical and topical. From Table~\ref{tab:human-eval} we observe that CHRT$_{12}$ achieves the lowest toxicity rating of only 0.027 with a minimal loss in linguistic quality of 0.024 points as compared to the base GPT-2. Low standard deviation in human annotation scores for CHRT$_{12}$ further strengthens our argument. Finally, it should be noted that all the entries in Table~\ref{tab:human-eval} marked with $^\ast$ have a p-value of greater than 0.05 for a pair-wise T-test with CHRT$_{12}$. Since all the baselines have a statistically insignificant difference in Topicality, we make no conclusion about the superiority of any approach as compared to ours.


\section{Limitations}
Our work is limited in capturing the unintended dependencies of attributes. It is possible that maximizing certain attributes like positive sentiment may maximize attributes like gender bias. A formal study to capture the dependency of the bias with varied attribute control is an important future direction. The efficacy automated metrics used to measure the linguistic qualities and attribute alignment of the generations is limited ~\cite{jozefowicz2016exploring}. Devising more exhaustive and explainable metrics is also an important future-work.

\section{Conclusion}
We present CHRT, a learning-based controlled language generation framework that achieves state-of-the-art attribute control while minimizing the loss in linguistic quality as compared to five recent baselines. The ability to combine control over multiple attributes and the ultra-fast inference of our approach makes it ideal for latency-constrained use-cases. We empirically showcase the effectiveness of our approach by performing both large-scale automated and human-evaluated benchmarks. For future work, we will like to work on making our approach more plug-and-play and achieve attribute control with an even lower loss in linguistic quality.

\bibliography{anthology,acl2020}
\bibliographystyle{acl_natbib}
\appendix
\section{CHRT Ablation Study}~\label{app:choice}
We perform ablation studies to justify the transformation block design choices.
\begin{table*}[h!]
\centering
\scalebox{0.85}{
    \begin{tabular}{l|cc|c|ccc}
    \hline
    \multicolumn{1}{c|}{\multirow{2}{*}{\textbf{\# Blocks}}} &
      \multicolumn{2}{c|}{\textbf{Toxicity}} &
      \multicolumn{1}{c|}{\textbf{Fluency}} &
      \multicolumn{3}{c}{\textbf{Diversity}} \\
    \multicolumn{1}{c|}{} &
      \multicolumn{1}{c}{\textbf{Avg. Max. Toxicity ($\downarrow$)}} &
      \multicolumn{1}{c|}{\textbf{Avg. Toxicity Prob. ($\downarrow$)}} &
      \multicolumn{1}{c|}{\textbf{Perplexity ($\downarrow$)}} &
      \multicolumn{1}{c}{\textbf{Dist-1 ($\uparrow$)}} &
      \multicolumn{1}{c}{\textbf{Dist-2 ($\uparrow$)}} &
      \multicolumn{1}{c}{\textbf{Dist-3 ($\uparrow$)}} \\
      \hline
    \textbf{1}       & \multicolumn{1}{r|}{0.210} & 0.075 & \textbf{20.271} & \multicolumn{1}{r|}{0.536} & \multicolumn{1}{r|}{0.817} & 0.814 \\
    \textbf{2}         & \multicolumn{1}{r|}{0.182} & 0.070 & 21.117 & \multicolumn{1}{r|}{\textbf{0.542}} & \multicolumn{1}{r|}{0.819} & 0.815 \\
    \textbf{3}        & \multicolumn{1}{r|}{0.170} & 0.071 & 21.116 & \multicolumn{1}{r|}{0.539} & \multicolumn{1}{r|}{\textbf{0.820}} & \textbf{0.817} \\
    \textbf{4}        & \multicolumn{1}{r|}{0.166} & 0.069 & 21.181 & \multicolumn{1}{r|}{0.540} & \multicolumn{1}{r|}{0.819} & 0.815 \\
    \textbf{5}        & \multicolumn{1}{r|}{\textbf{0.160}} & \textbf{0.068} & 21.404 & \multicolumn{1}{r|}{0.540} & \multicolumn{1}{r|}{\textbf{0.820}} & 0.816 \\
    \hline
    \end{tabular}
}
\caption{Ablation Study for justifying the number of identical smaller blocks in CHRT.}
\label{tab:res-ablation_nblocks}
\end{table*}
\begin{table*}[h!]
\centering
\scalebox{0.85}{
\begin{tabular}{l|cc|c|ccc}
\hline
\multicolumn{1}{c|}{\multirow{2}{*}{\textbf{$\kappa$}}} &
  \multicolumn{2}{c|}{\textbf{Toxicity}} &
  \multicolumn{1}{c|}{\textbf{Fluency}} &
  \multicolumn{3}{c}{\textbf{Diversity}} \\
\multicolumn{1}{c|}{} &
  \multicolumn{1}{c}{\textbf{Avg. Max. Toxicity ($\downarrow$)}} &
  \multicolumn{1}{c|}{\textbf{Avg. Toxicity Prob. ($\downarrow$)}} &
  \multicolumn{1}{c|}{\textbf{Perplexity ($\downarrow$)}} &
  \multicolumn{1}{c}{\textbf{Dist-1 ($\uparrow$)}} &
  \multicolumn{1}{c}{\textbf{Dist-2 ($\uparrow$)}} &
  \multicolumn{1}{c}{\textbf{Dist-3 ($\uparrow$)}} \\
  \hline
\textbf{0.5}       & \multicolumn{1}{r|}{0.210} & 0.075 & \textbf{20.271} & \multicolumn{1}{r|}{0.536} & \multicolumn{1}{r|}{0.817} & 0.814 \\
\textbf{1}         & \multicolumn{1}{r|}{0.181} & 0.072 & 20.802 & \multicolumn{1}{r|}{0.540} & \multicolumn{1}{r|}{0.819} & \textbf{0.815} \\
\textbf{2}        & \multicolumn{1}{r|}{\textbf{0.166}} & \textbf{0.070} & 21.159 & \multicolumn{1}{r|}{\textbf{0.546}} & \multicolumn{1}{r|}{\textbf{0.820}} & \textbf{0.815} \\
\hline
\end{tabular}}
\caption{Ablation Study for justifying the choice of $\kappa$.}
\label{tab:res-ablation_hdim}
\end{table*}
\subsection{Number of blocks}
As shown in Figure ~\ref{fig:sub3}, our transformation block is a concatenation of two identical blocks. In this ablation study we vary the number of identical blocks. We perform the same experiment as in Section ~\ref{ssec:toxicity} to evaluate attribute control of toxicity and linguistic quality of generation. For our models, we take CHRT$_{12}$ with varying number of blocks. Table ~\ref{tab:res-ablation_nblocks} summarizes this ablation study. We can observe that more the number of blocks, the better performance we get in terms of both reducing perplexity and increasing attribute control. However, the gain is minimal after more than 2 blocks thus justifying our choice of having 2 identical blocks in our transformation.
\subsection{Intermediate Hidden Dimension}\label{app:loss_weights}
The constituent blocks of our transformation block have linear layers with the intermediate dimension of $\kappa*h$ where $\kappa \in \mathbb{R}$ is the multiplication factor and $h$ is the hidden layer dimension of the base model. As shown in Figure ~\ref{fig:sub3}, we have selected $\kappa$ to be 0.5. We justify this choice empirically through a similar ablation study as for \#blocks. Table ~\ref{tab:res-ablation_hdim} shows that for $\kappa = 0.5$, we have the highest fluency (lowest perplexity) with a minor loss in attribute control (toxicity). Further, the number of weights for selecting $\kappa=0.5$ as compared to 1 or 2 is drastically less, leading to faster training and inference. Therefore, we select $\kappa$ to be 0.5.

\section{Training and Generation Details}~\label{app:baselines}
All training and inference are performed on a machine with 8 Tesla V100 GPUs using the Hugginface and PyTorch library. For the baselines, we directly run their publicly available code without modifications.
\subsection{GPT-2}
GPT-2 Medium with 355 million parameters is used as our base model in the paper. For generation, we use huggingface's generate function with the following hyper-parameters:

\begin{enumerate}
    \item \textbf{No. of Generations:} 25 \vspace{-10pt}
    \item \textbf{Max-Length:} 25 \vspace{-10pt}
    \item \textbf{Top-p:} 0.8 \vspace{-10pt}
    \item \textbf{Repetition Penalty: } 1.2 \vspace{-10pt}
\end{enumerate}

\subsection{NLC}
For NLC, we use huggingface library's $bad\_words\_ids$ parameter to negatively constrain the model over a bag of word. We do a search for PMI threshold $\theta$ to create the bag of bad words. We select the one with highest sentiment control and lowest loss in generation quality. The best size of negative words is 123,950, 94,83, and 74,502 for simplicity, toxicity and sentiment control respectively. Rest of the generation parameters are same as GPT-2.

\subsection{DAPT}
We fine-tune DAPT on positive attribute class for our experiments. This includes non-toxic text as extracted from RealToxicityPrompts, positive sentiment text and simple text from RealSentimentPrompts and RealSimplicityPrompts respectively. The details for these datasets can be found in Appendix. We again use huggingface library's generate function with same parameters as for GPT-2.

\subsection{PPLM}
We train toxicity, sentiment and simplicity classifiers using the original authors public code. All the hyper-parameters are preserved as in the original paper. For a fair comparison, we retrain these classifiers on the same data as rest of the baselines using the original author's released codebase~\footnote{https://github.com/uber-research/PPLM/}.

\subsection{GeDi}
Similar to PPLM, we retrain our own GeDi models for toxicity, sentiment and simplicity on same datasets using the code released by authors~\footnote{https://github.com/salesforce/GeDi}. All the training and generation hyper-parameters are preserved from the original paper and the official released code is used without modifications.

\subsection{DExperts}
The experts and dexperts models are retrained using the author's puiblicly released code~\footnote{https://github.com/alisawuffles/DExperts}. For generation, the authors suggest varying the $\alpha$ parameters which controls the strength of generation. For each experiment, we vary $\alpha$ from 1 to 4 in steps of 0.2 and select the best $\alpha$ in terms of attribute control with minimal loss in perplexity. We select $\alpha$ to be 2, 3.2 and 3.0 for detoxification, sentiment control and simplicity attribute control experiments respectively because the gain in attribute control was minimal as compared to loss in perplexity above these values. Finally, we generate 25 generations of length 25 tokens each, with the Top-p for nucleus sampling being 0.9, the value authors suggested.

\subsection{CHRT}
For fine-tuning the guider models $\mathcal{LM}^+$ and $\mathcal{LM}^-$ we use the huggingface's fine-tuning scripts with the difference that we lock the language-modelling head layer weights, as explained in Section ~\ref{ssec:guider}. Following are the hyperparameter details for fine-tuning, all of them are the default parameters from the huggingface finetuning script:

\begin{enumerate}
    \item \textbf{Model:} gpt2-medium\vspace{-5pt}
    \item \textbf{No. of Parameters:} 345M\vspace{-5pt}
    \item \textbf{No. of Epochs:} 3\vspace{-5pt}
    \item \textbf{Block Size:} 128\vspace{-5pt}
    \item \textbf{No. of Gradient Accumulation Steps:} 16\vspace{-5pt}
    \item \textbf{Optimizer:} AdamW\vspace{-5pt}
    \item \textbf{Learning Rate:} 2e-5\vspace{-5pt}
    \item \textbf{Learning Rate:} 2e-5\vspace{-5pt}
    \item \textbf{Effective Batch Size:} 32\vspace{-5pt}
\end{enumerate}

For training the transformation head, we use the following hyperparameters:
\begin{enumerate}
    \item \textbf{No. of Epochs:} 3\vspace{-5pt}
    \item \textbf{Effective Batch Size:} 16\vspace{-5pt}
    \item \textbf{Optimizer:} AdamW\vspace{-5pt}
    \item \textbf{Learning Rate:} 2e-5\vspace{-5pt}
    \item \textbf{Triplet Loss Margin:} 1\vspace{-5pt}
    \item \textbf{Triplet Loss p:} 2\vspace{-5pt}
\end{enumerate}

Finally, for generation we use the following hyperparameters:

\begin{enumerate}
    \item \textbf{No. of Generations:} 25 \vspace{-5pt}
    \item \textbf{Max-Length:} 25 \vspace{-5pt}
    \item \textbf{Top-p:} 0.8 \vspace{-5pt}
    \item \textbf{Repetition Penalty: } 1.2 \vspace{-5pt}
\end{enumerate}

\newpage
\section{Generation Examples}
We show example continuation generation for a randomly selected prompt for all the baselines. Table~\ref{tab:gens-toxic} and Table~\ref{tab:gens-sentiment} shows example continuations for the attribute of toxicity and sentiment.

\begin{table*}[]
\begin{tabularx}{\linewidth}{l X}
\hline
\multicolumn{1}{l}{\textbf{Model}} & \multicolumn{1}{c}{\textbf{Continuation}}  \\ \hline
\textbf{GPT-2}       & shit. This definition makes it sound like every writer has the same syndrome, and when you read what they say on there site... \\ \hline
\textbf{NLC} &
  gibberish, or in some cases vulgarity. In one version of the story above (from 2008), they wrote... \\ \hline
\textbf{DAPT} &
  shit. It's an idea the author of The Real Housewives, Kim Richards has embraced on social media and elsewhere it... \\ \hline
\textbf{PPLM} &
  the use and and in the content ( of content and, which can be a condition or condition, mental state or state and... \\ \hline
\textbf{GeDi} &
  be of little or no importance or value. Ching Chong's definition is no better than what I think... \\ \hline
\textbf{DExperts} &
  to be more like it is (The Economist, June 25, 1997, p. 71) garbage, Webster Dictionary defines it... \\ \hline
\textbf{CHRT$_{21}$} &
  be full of "emotional The patient may say he can't find the words, but it is evident when his writings... \\ \hline
\textbf{CHRT$_{11}$} & music. In the past few years we have seen several cases in which individuals began using software applications for...        \\ \hline
\textbf{CHRT$_{12}$} &
  be about what they would rather not - He says, "I need it for the family... \\ \hline
\end{tabularx}
\caption{\textbf{Prompt}: Webster’s Dictionary defines Shit Writing Syndrome as “a disorder that turns one’s writing to...}
\label{tab:gens-toxic}
\end{table*}

\begin{table*}[]
\begin{tabularx}{\linewidth}{l X}
\hline
\multicolumn{1}{l}{\textbf{Model}} & \multicolumn{1}{c}{\textbf{Continuation}}  \\ \hline
\textbf{GPT-2}       & divorce. He would say all sorts a lot but never had any success in doing it because he was not serious about me having... \\ \hline
\textbf{NLC} & brother. My sister has a history with her ex-husband's girlfriend who she thought was an angel but is actually just another... \\ \hline
\textbf{DAPT} & death The best part is that he was really happy about this: Its a great opportunity to make sure kids... \\ \hline
\textbf{PPLM} & wife, who, for some reason, is the last person of some years ago I have to this year. I have never... \\ \hline
\textbf{GeDi} & divorce. So I chose to leave. He's still my best friend. Anyway, last year he... \\ \hline
\textbf{DExperts} & abrupt passing. I love him to pieces (at least to a degree). Thank you to my long list gedi... \\ \hline
\textbf{CHRT$_{21}$} & marriage to Myra Lynn.  My first thought was thank God that I had a wonderful supportive family with great friends in... \\ \hline
\textbf{CHRT$_{11}$} &  amazing work as an incredible mentor I love him so much. I really admire how open he was about each project for us...        \\ \hline
\textbf{CHRT$_{12}$} & wonderful team!!! Love you all!!!! :) Thanks for sharing your memories!! Thank You from everyones hearts Hope this helps... \\ \hline
\end{tabularx}
\caption{\textbf{Prompt}: Sadly, they were honestly the worst five years of my life! Why? Because of the emotional, financial and legal hell created by his...}
\label{tab:gens-sentiment}
\end{table*}

\end{document}